\documentclass[10pt,twocolumn,letterpaper]{article}

\usepackage[pagenumbers]{cvpr} 

\usepackage{graphicx}
\usepackage{amsmath}
\usepackage{amssymb}
\usepackage{booktabs}

\usepackage{bm}
\usepackage{xcolor, colortbl}
\usepackage{url}
\usepackage{comment}

\usepackage{sidecap}
\sidecaptionvpos{figure}{t} 

\definecolor{green}{rgb}{0.0, 0.5, 0.0}
\definecolor{red}{rgb}{0.9, 0.0, 0.0}
\definecolor{blue}{rgb}{0.0, 0.0, 0.9}

\newcommand{\tableCellHeight}{1}
\newcommand{\tabstyle}[1]{
  \setlength{\tabcolsep}{#1}
  \renewcommand{\arraystretch}{\tableCellHeight}
  \centering
  \small
}

\usepackage{blindtext}
\usepackage[export]{adjustbox} 
\usepackage{subfig} 
\usepackage{wrapfig}

\usepackage{tabularx}
\newcolumntype{C}{>{\centering\arraybackslash}X}
\usepackage{tabu}

\renewcommand{\paragraph}[1]{\vspace{1mm}\noindent\textbf{#1}.}

\usepackage[pagebackref,breaklinks,colorlinks]{hyperref}

\usepackage[capitalize]{cleveref}
\crefname{section}{Sec.}{Secs.}
\Crefname{section}{Section}{Sections}
\Crefname{table}{Table}{Tables}
\crefname{table}{Tab.}{Tabs.}

\def\cvprPaperID{****} 
\def\confName{CVPR}
\def\confYear{2023}

\begin{document}

\title{On-Device Domain Generalization}

\author{Kaiyang Zhou, Yuanhan Zhang, Yuhang Zang, Jingkang Yang, Chen Change Loy, Ziwei Liu\thanks{Corresponding author}\\
S-Lab, Nanyang Technological University, Singapore
}
\maketitle

\begin{abstract}
We present a systematic study of domain generalization (DG) for tiny neural networks. This problem is critical to on-device machine learning applications but has been overlooked in the literature where research has been merely focused on large models. Tiny neural networks have much fewer parameters and lower complexity and therefore should not be trained the same way as their large counterparts for DG applications. By conducting extensive experiments, we find that knowledge distillation (KD), a well-known technique for model compression, is much better for tackling the on-device DG problem than conventional DG methods. Another interesting observation is that the teacher-student gap on out-of-distribution data is bigger than that on in-distribution data, which highlights the capacity mismatch issue as well as the shortcoming of KD. We further propose a method called out-of-distribution knowledge distillation (OKD) where the idea is to teach the student how the teacher handles out-of-distribution data synthesized via disruptive data augmentation. Without adding any extra parameter to the model---hence keeping the deployment cost unchanged---OKD significantly improves DG performance for tiny neural networks in a variety of on-device DG scenarios for image and speech applications. We also contribute a scalable approach for synthesizing visual domain shifts, along with a new suite of DG datasets to complement existing testbeds.
\end{abstract}

\section{Introduction}
\label{sec:intro}

\begin{table*}[t]
	\tabstyle{21pt}
	\caption{Large vs.~tiny neural networks. Due to capacity mismatch, the three tiny neural networks perform worse than their large counterpart, especially on OOD data---the ID performance gap is around 12\% while the OOD performance gap is around 16\%.}
	\label{tab:model_specs}
	\begin{tabular}{l ccccc}
		\toprule
		Model & Params & Size & MACs & ID Acc & OOD Acc \\
		\midrule
		ResNet50 & 25.56M & 97.70MB & 4133.74M & 77.4\% & 60.2\% \\
		MobileNetV3-Small & 1.50M & 5.79MB & 61.44M & 64.0\% & 42.9\%  \\
		MobileNetV2-Tiny & 0.75M & 2.91MB & 65.20M & 63.6\% & 42.0\% \\
		MCUNet & 0.74M & 2.87MB & 145.13M & 66.8\% & 46.4\% \\
		\bottomrule
	\end{tabular}
	\begin{flushleft}
		\footnotesize
		\itshape
		ID means in-distribution. OOD means out-of-distribution. Acc means accuracy (average performance on the newly-proposed DOSCO-2k benchmark).
	\end{flushleft}
\end{table*}

\begin{figure*}[t]
	\centering
	\subfloat[Improvement over ERM.]{
		\includegraphics[width=.32\textwidth, valign=t]{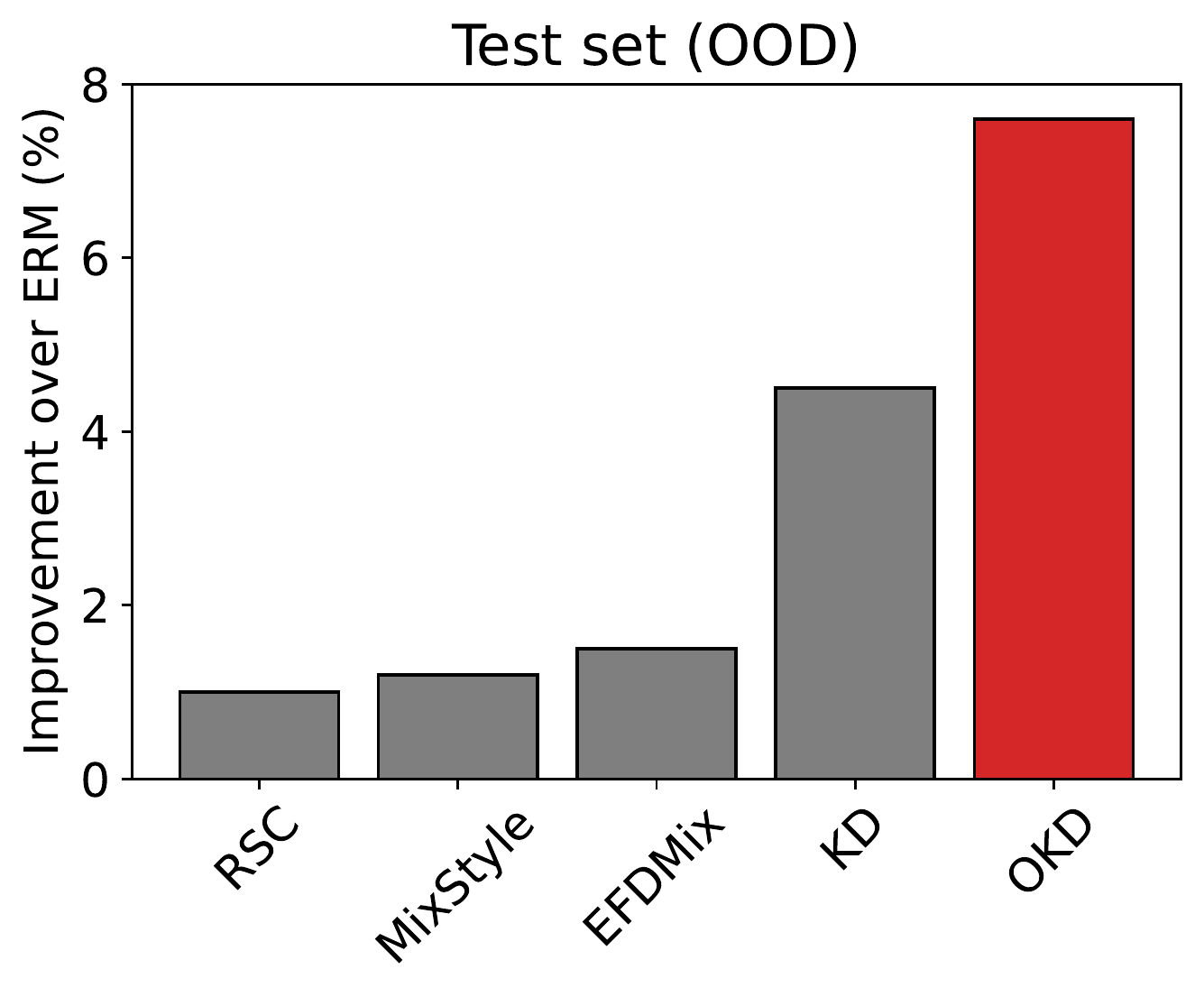}
		\label{fig:teaser_mbv3small}
	}
	~
	\subfloat[The teacher-student gap on out-of-distribution (OOD) data.]{
		\includegraphics[width=.32\textwidth, valign=t]{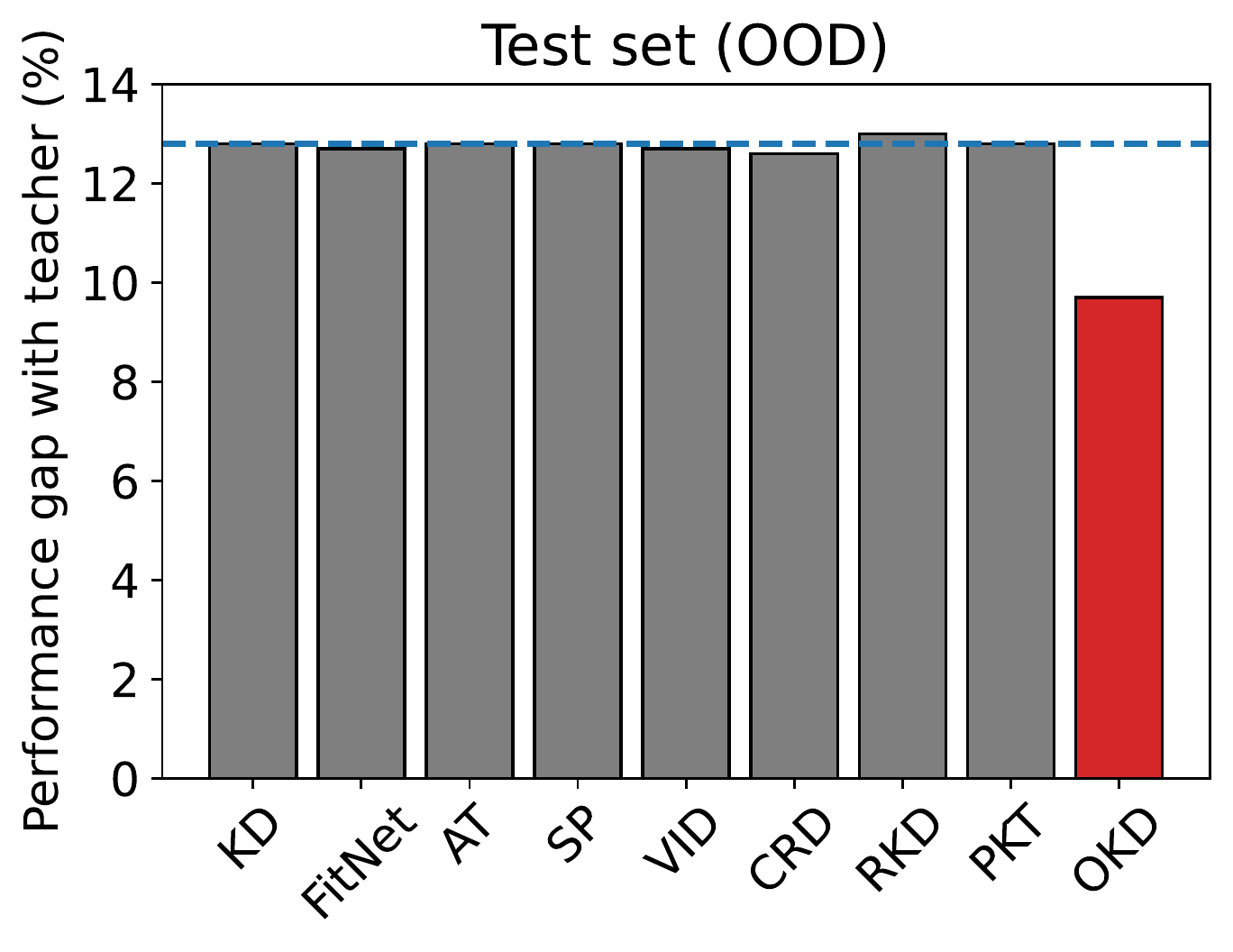}
		\vphantom{\includegraphics[width=.32\textwidth, valign=t]{images/teaser_mbv3small.pdf}} 
		\label{fig:student_vs_teacher_ood}
	}
	~
	\subfloat[The teacher-student gap on in-distribution (ID) data.]{
		\includegraphics[width=.32\textwidth, valign=t]{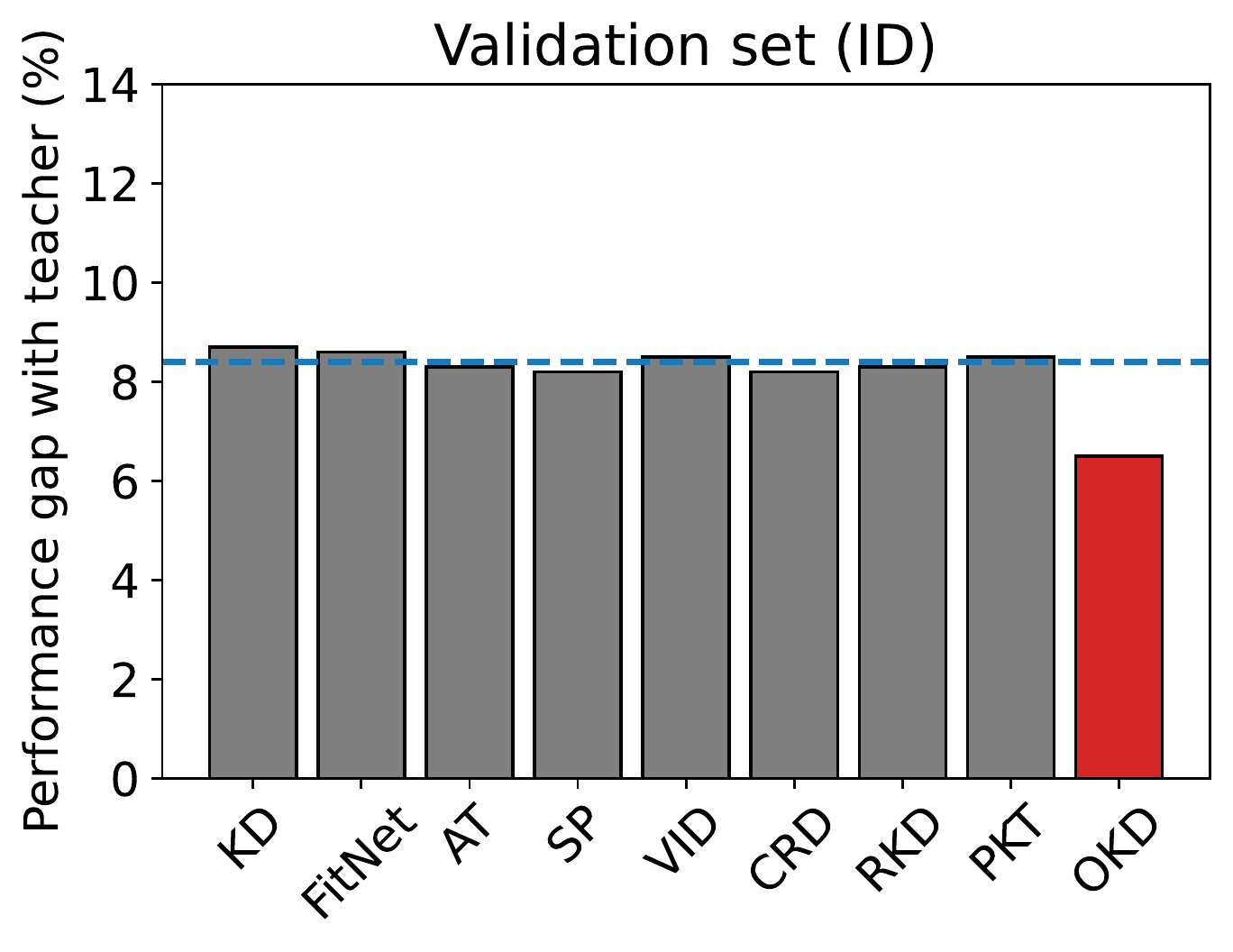}
		\vphantom{\includegraphics[width=.32\textwidth, valign=t]{images/teaser_mbv3small.pdf}} 
		\label{fig:student_vs_teacher_id}
	}
	\caption{Overview of the main results on our new DOSCO-2k benchmark using tiny neural networks. OKD achieves significant improvements over the DG and KD-based methods, as shown in (a) and (b). Notably, OKD's gain on OOD performance is more significant than its gain on ID performance, evidenced by the reduced gaps in (b) and (c). The blue dashed lines in (b) and (c) denote the average performance of all KD-based methods.}
	\label{fig:overview}
\end{figure*}

Domain generalization (DG), also known as out-of-distribution (OOD) generalization, is a problem concerned with whether or not a model learned from source data can perform well on unseen target data with domain shifts~\cite{blanchard2011generalizing,muandet2013domain}. In the last decade, DG has been extensively studied in the literature~\cite{zhou2022domain}, especially for neural networks that have become the mainstream approach in machine learning and pattern recognition tasks. However, existing DG research mainly focuses on large models with massive parameter sizes and heavy computations.

This paper, for the first time, presents a systematic study on methods that can improve DG for \textit{tiny neural networks}. Tiny neural networks are critical to on-device machine learning applications~\cite{cai2022enable}, which have received increasing attention due to the rapid increase in low-cost mobile devices, such as mobile phones and IoT devices. However, running neural networks on low-power devices is challenging due to strict requirements on model size (storage) and latency. For example, IoT devices with microcontroller units (MCUs) typically have an SRAM smaller than 512KB, which is too small to fit most neural networks~\cite{lin2021mcunetv2}. See Table~\ref{tab:model_specs} for a comparison of model specifications between neural networks of different sizes.

DG is essential for tiny neural networks because mobile devices are often used in an unanticipated environment, suggesting that the model must have the ability to overcome any domain shift. However, making tiny neural networks domain-generalizable is nontrivial since they have much fewer parameters than large neural networks, and hence much smaller capacity. As shown in Table~\ref{tab:model_specs}, the performance declines significantly for tiny models, especially on OOD data.

Due to the differences, tiny neural networks should not be trained the same way as their large counterparts for DG applications. We observe that state-of-the-art DG methods only bring a limited improvement for tiny models, at most one or two percent increases in accuracy over Empirical Risk Minimization (ERM) that is known as a strong baseline method~\cite{gulrajani2020search}. See the results of RSC~\cite{huang2020self}, MixStyle~\cite{zhou2021domain} and EFDMix~\cite{zhang2022exact} in Figure~\ref{fig:teaser_mbv3small}.

A straightforward solution is to use knowledge distillation (KD)~\cite{bucilua2006model,hinton2015distilling}, which is a well-known technique for compressing large models into smaller ones. We indeed find that KD boosts DG performance more significantly than DG methods, as shown in Figure~\ref{fig:teaser_mbv3small}. Nevertheless, the teacher-student performance gap on OOD test data is still huge. Such a gap exists for a wide range of KD-based methods, as evidenced in Figure~\ref{fig:student_vs_teacher_ood}. Interestingly, we observe that the gap on OOD data is much bigger than that on in-distribution data (see Figure~\ref{fig:student_vs_teacher_ood} and~\ref{fig:student_vs_teacher_id}). The results suggest that the capacity mismatch issue makes it harder to distill generalizable knowledge.

We argue that the reason why the student failed to match the teacher's OOD performance is because the student was never taught how to handle OOD data---the conventional KD loss is computed on in-distribution data only. To mitigate the problem, we propose \textit{out-of-distribution knowledge distillation} (OKD), a simple method that extends KD with a new distillation loss computed on OOD data. Since collecting extra OOD data would be challenging and costly for downstream datasets, we resort to using disruptive data augmentation methods to synthesize OOD data: the main idea is to make input significantly diverge from the support of the training data distribution but not completely disjoint. Despite the simplicity, OKD significantly improves upon KD without increasing the deployment cost (see Figure~\ref{fig:overview} for an overview of OKD's results).

We also contribute a scalable and extremely simple approach for synthesizing visual domain shifts, which can be applied to broad data (i.e., any image dataset) at scale without much human effort. Specifically, we observe that common visual domain shifts are closely related to contextual shifts and use a neural network trained on the Places dataset~\cite{zhou2017places} to automatically cluster visual contexts in images. This is also inspired by a prior research finding~\cite{zhou2014object} that Places-trained models can extract meaningful patterns associated with the composition of scenes, namely visual contexts. By applying this approach to 7 image datasets, we build a new benchmark named \textit{DOmain Shifts in COntext} (DOSCO), which can complement existing testbeds. Compared with existing datasets, our datasets cover wider categories including generic objects, fine-grained categories like aircrafts and animals, and human actions, and contain a broader spectrum of domain shifts, such as shifts in image style, background, viewpoint, object pose, and so on.

The main contributions and findings of this paper are summarized as follows. (1) For the first time, the problem of on-device DG is systematically studied, giving the community timely insights on how to improve DG performance for tiny, low-power models. (2) Comprehensive results are presented to demonstrate KD's advantages over conventional DG methods as well as its shortcoming. (3) A simple extension to KD is proposed, which achieves significant improvements in a variety of on-device DG scenarios for image and speech applications. (4) A scalable approach for synthesizing visual domain shifts is proposed, along with a new suite of DG datasets, which, compared with existing datasets, contain wider categories and broader domain shifts. The source code, models and datasets are released to facilitate future research on on-device DG.\footnote{\url{https://github.com/KaiyangZhou/on-device-dg}}

\section{Approach}
\label{sec:approach}

\begin{figure}[t]
    \centering
    \includegraphics[width=\columnwidth]{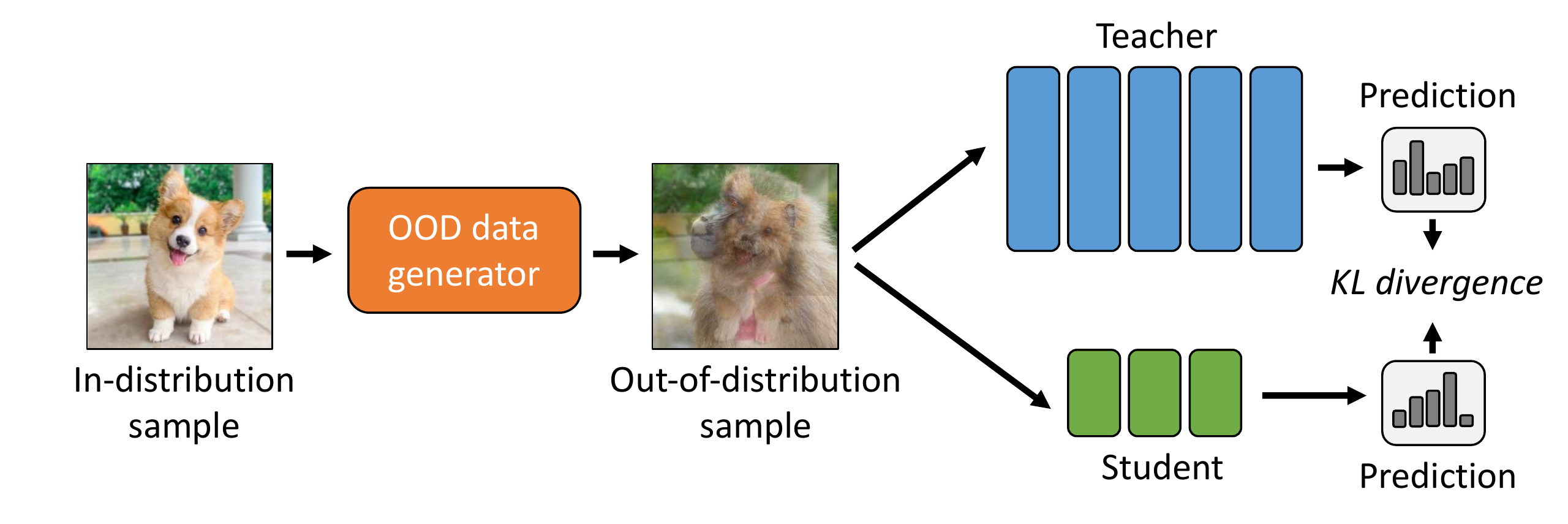}
    \caption{\textbf{OKD architecture}. The main idea is to teach the student how the teacher handles OOD data, which is synthesized using disruptive data augmentation methods.}
    \label{fig:okd_framework}
\end{figure}

We design a frustratingly simple approach to tackle on-device DG, named out-of-distribution knowledge distillation (OKD), which essentially extends knowledge distillation (KD) with a new distillation loss computed on synthetic OOD data. Figure~\ref{fig:okd_framework} illustrates the idea.

\subsection{Background on Knowledge Distillation}
\label{sec:approach;subsec:kd}

The main idea of KD is to use one or multiple (ensemble) big models, called \textit{teacher}, to guide the learning of a small model, called \textit{student}---the latter is more suitable to be deployed on low-power devices, such as mobile phones or IoT devices. KD uses an auxiliary loss function (typically a distance measure) that encourages the student's output to mimic the teacher's output, along with a task-related loss, such as the cross-entropy for classification.

Formally, let $\bm{x}$ denote an input (e.g., an image or an audio waveform) and $y$ the label, the overall loss function for KD can be written as
\begin{equation} \label{eq:kd}
	L_{KD} = \lambda H( y, f_S(\bm{x}) ) + (1-\lambda) D_{KL}( f_S(\bm{x}), f_T(\bm{x}) ),
\end{equation}
where the first term is the cross-entropy loss, and the second term is the KL divergence between the outputs of the student $f_S$ and teacher $f_T$. $\lambda$ is a balancing weight, which is often set to 0.1 according to the KD literature~\cite{tian2020contrastive}.

The prediction probability $\bm{p}$ for the input $\bm{x}$ is computed as
\begin{equation} \label{eq:prob}
	\bm{p}_i = \frac{\exp(\bm{z}_i/\pi)}{\sum_j \exp(\bm{z}_j/\pi)},
\end{equation}
where $\bm{z}$ denotes logits, and $\pi$ is a temperature parameter. The common practice is to set $\pi=1$ for the first term in Eq.~\ref{eq:kd} and $\pi=4$ for the second term to soften the probability distribution~\cite{tian2020contrastive}.

\subsection{Out-of-Distribution Knowledge Distillation}
\label{sec:approach;subsec:okd}

The key idea of OKD is to teach the student how the teacher handles OOD data. This objective is vital for distilling generalizable knowledge from the teacher to the student but is missing in the conventional KD formulation.

Let $A(\cdot)$ denote a data augmentation function, called OOD data generator, which aims to make the input deviate from the support of the source data distribution. The formulation of OKD is as follows:
\begin{equation} \label{eq:okd}
	L_{OKD} = L_{KD} + (1-\lambda) D_{KL}( f_S(A(\bm{x})), f_T(A(\bm{x})) ).
\end{equation}

Compared with KD in Eq.~\ref{eq:kd}, OKD just adds a new term for distilling the teacher's knowledge about OOD data, which is easy to implement. Moreover, OKD adds a negligible overhead during training while leaving the inference unchanged because the model architecture remains the same. Note that the new term has the same balancing weight as the KD term in Eq.~\ref{eq:kd}.

\paragraph{OOD data generator}
Since collecting extra OOD data for downstream datasets is impractical, we propose to use disruptive data augmentation methods. The main idea is to make input significantly diverge from the support of the training data distribution but not completely disjoint---we later show that using disjoint data harms the performance. In our experiments, we evaluate a list of candidate methods and provide users with empirical evidence and valuable insights on how to design the OOD data generator, for image and speech applications respectively. Notably, $A(\bm{x})$ does not need to maintain the semantics of $\bm{x}$, e.g., for image recognition we can use CutMix~\cite{yun2019cutmix} or Mixup~\cite{zhang2017mixup} to generate images not belonging to any existing class. In future work, it would be interesting to explore more sophisticated augmentation functions, such as making the function fully learnable.

\section{The Domain Shifts in Context Benchmark}
\label{sec:dosco}

 \begin{figure*}[t]
 	\centering
 	\subfloat[P-Air (class: 737-200).]{
 		\includegraphics[width=.31\textwidth, valign=t]{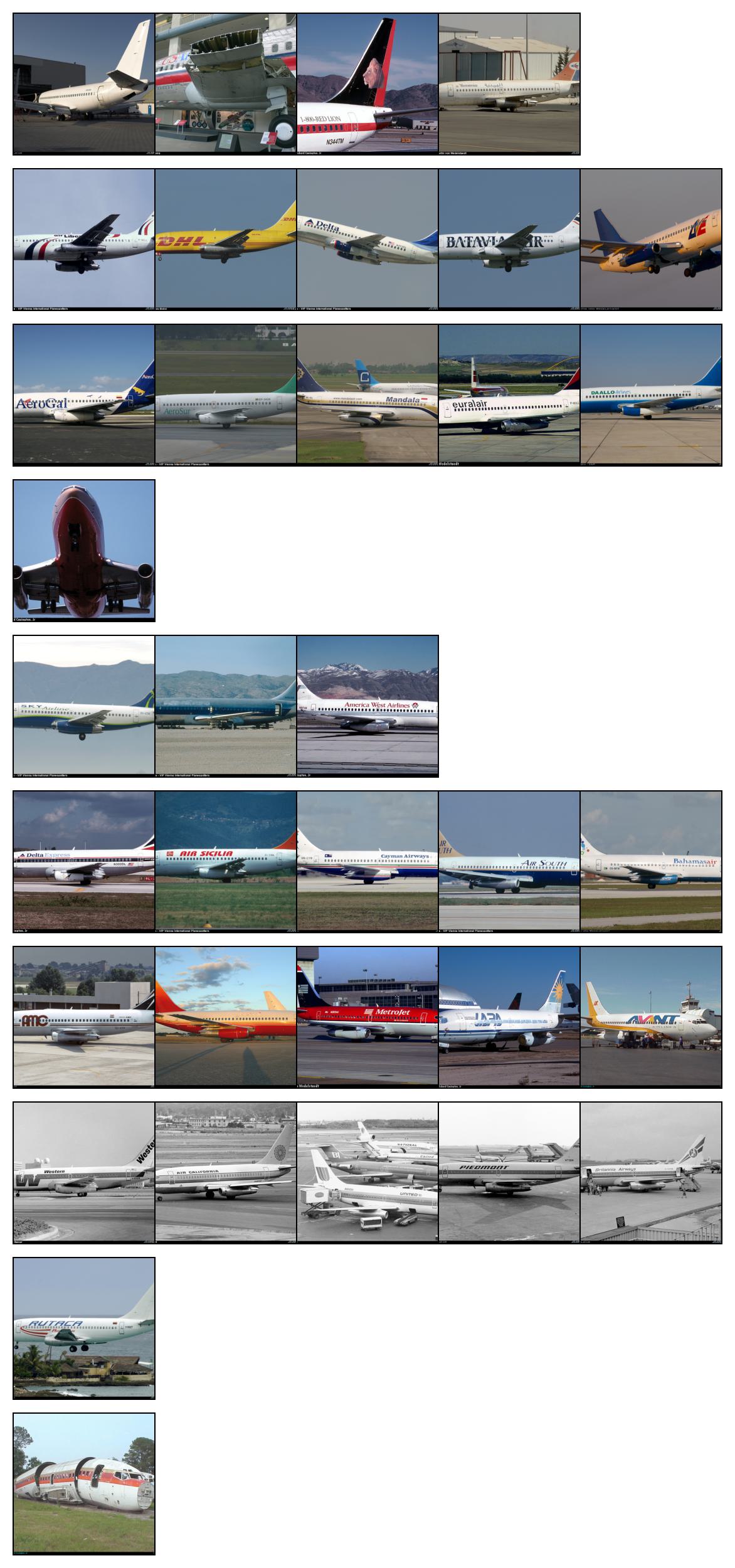}
 		\label{fig:737-200}
 	}
 	~
 	\subfloat[P-Ctech (class: crab).]{
 		\includegraphics[width=.31\textwidth, valign=t]{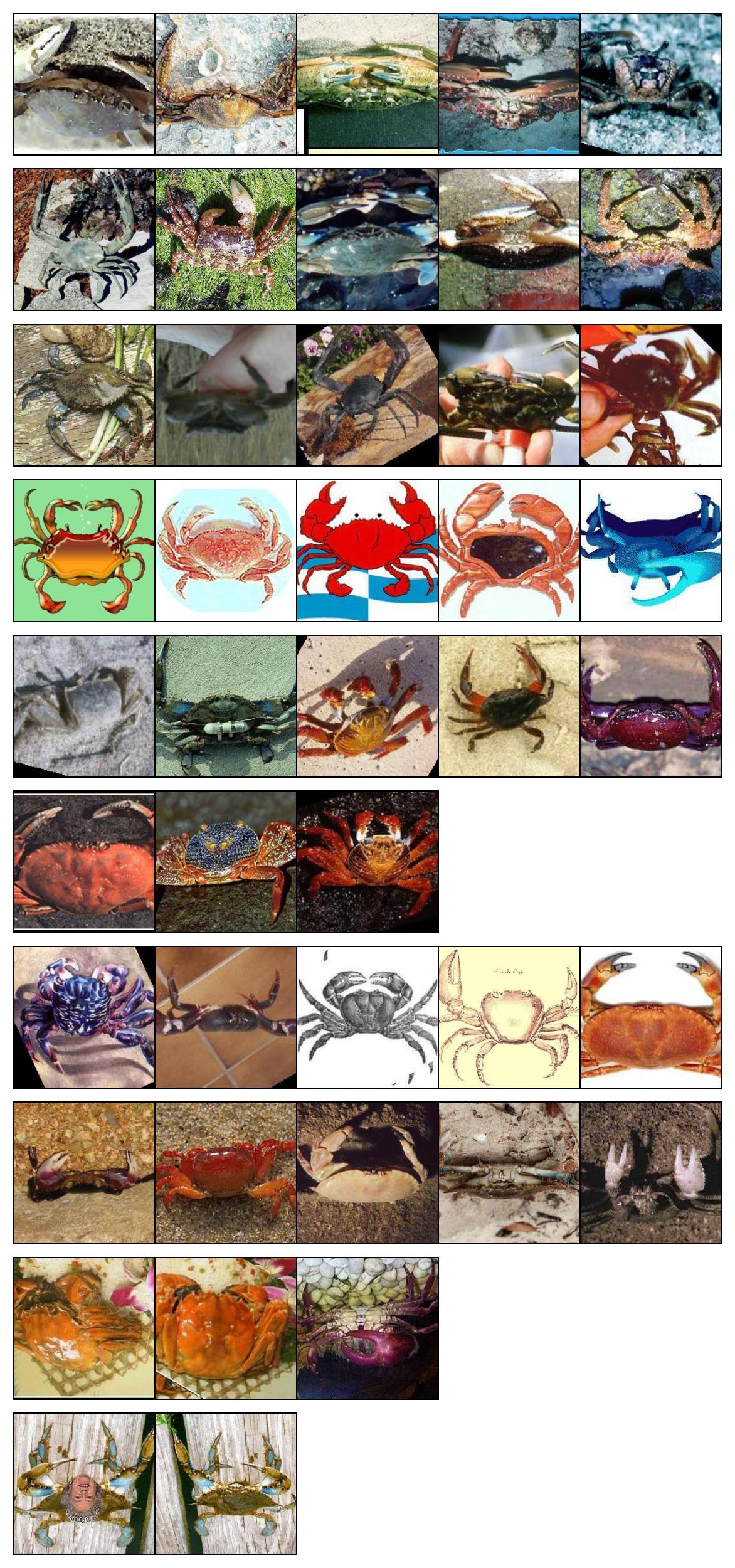}
 		\label{fig:crab}
 	}
 	~
 	\subfloat[P-Pets (class: pug).]{
 		\includegraphics[width=.31\textwidth, valign=t]{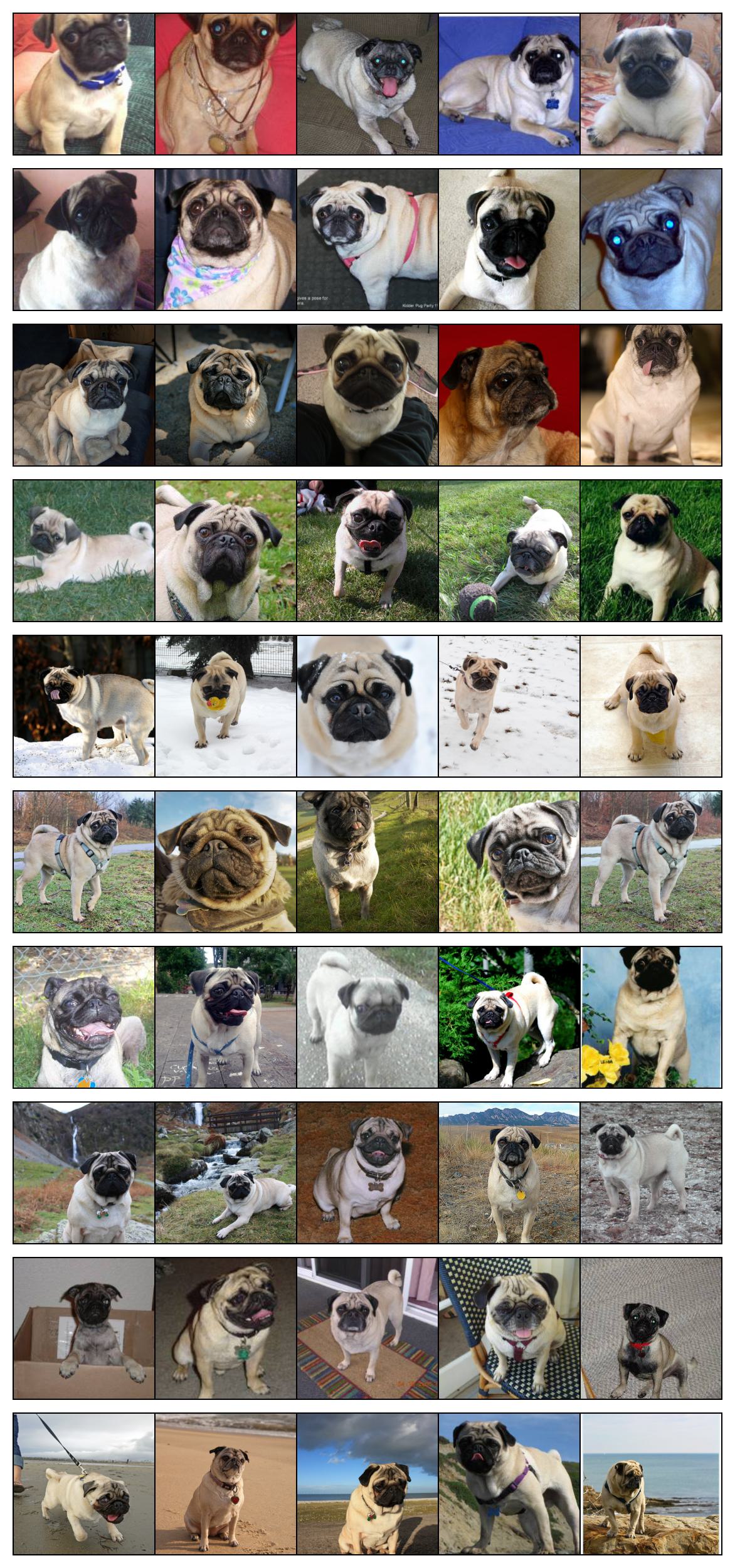}
 		\label{fig:pug}
 	}
 	\caption{Example images from three DOSCO datasets. Each row contains five random images with the same domain label (some have less than five images). It is worth noting that the domain shifts automatically discovered by a Places-learned model are \textit{highly diverse}, e.g., viewpoint and contrast changes in P-Air (a), style variations in P-Ctech (b), and background shifts in P-Pets (c).
 	}
 	\label{fig:few_dosco_examples}
 \end{figure*}

\begin{SCfigure*}
	\centering
	\includegraphics[width=.7\textwidth]{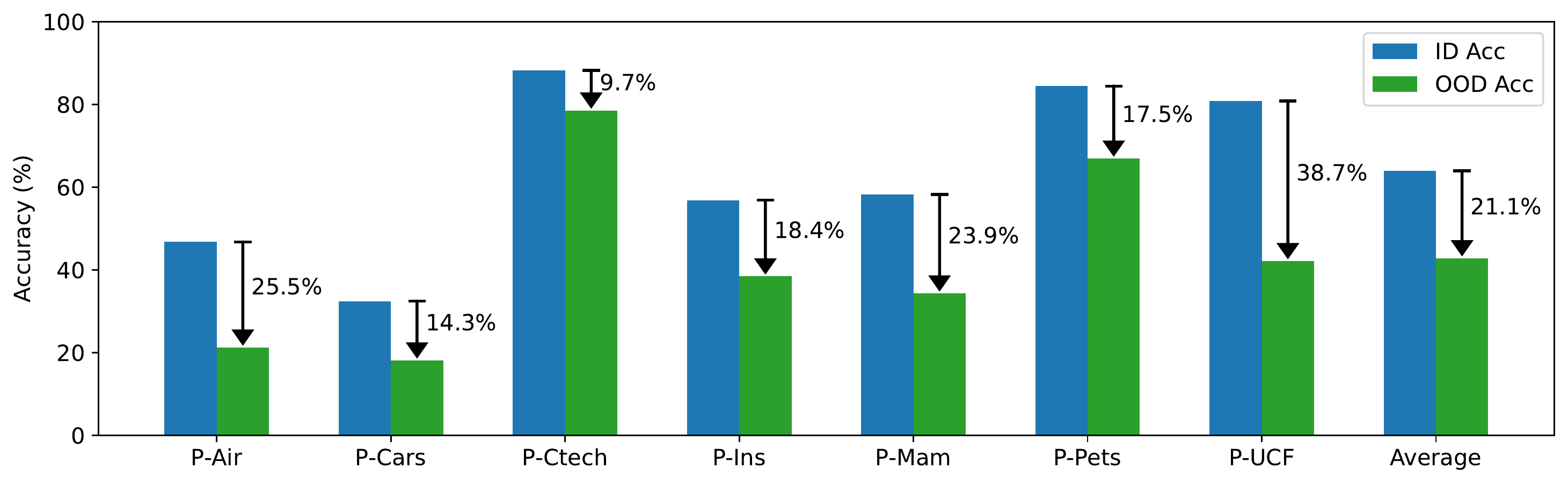}
	\caption{Comparison between ID and OOD performance on the DOSCO-2k benchmark. The big gaps highlight the domain shift problem. Example images of some datasets are shown in Figure~\ref{fig:few_dosco_examples}.}
	\label{fig:dosco_id_ood_gap}
\end{SCfigure*}

\paragraph{Motivation}
Building DG datasets from scratch is nontrivial: one needs to first define domain labels---which are often difficult to describe using natural language---and then use them to collect data from particular sources, e.g., the web. We aim to design a more scalable approach to build diverse DG datasets, which can complement existing testbeds.

\paragraph{Main idea}
In computer vision, domain shifts are often associated with visually perceptible changes in, for example, image style, background, viewpoint, contrast, object pose, and camera setups, to name a few. These domain shifts can essentially be summarized as a shift in \textit{visual context}. Based on this idea, we design a simple approach called DOmain Shifts in COntext (DOSCO) to automatically identify contextual information in images and create DG testbeds in an efficient way. Specifically, the DOSCO approach first trains a neural network on the Places dataset~\cite{zhou2017places} to extract contextual information in images, i.e., the composition of scenes that encapsulates all relevant image attributes. Then, images are clustered using features extracted by the neural network to synthesize domain labels.

\paragraph{Implementation}
We first fine-tune a ViT-Large model~\cite{dosovitskiy2020image,he2022masked} on Places365~\cite{zhou2017places}, which is dubbed \textit{PlacesViT}. Then, 7 image datasets commonly used in transfer learning research~\cite{zhai2019large,zhou2022learning} are chosen to be included in our benchmark: FGVCAircraft~\cite{maji2013fine}, StanfordCars~\cite{krause20133d}, Caltech101~\cite{fei2004learning}, Omni-Instrumentality~\cite{zhang2022benchmarking}, Omni-Mammal~\cite{zhang2022benchmarking}, OxfordPets~\cite{parkhi2012cats}, and UCF101~\cite{soomro2012ucf101}. For each dataset, the images within each class are clustered using K-means ($K\!=\!10$) based on the PlacesViT features. For each class, a random 50/50 split on the domain labels is then performed to produce a training set and a test set. A held-out validation set is randomly sampled from within the training set using a $2\!:\!8$ ratio. Each dataset has three random splits. Figure~\ref{fig:few_dosco_examples} shows some random images from three DOSCO datasets. More can be found in the supplementary.

\paragraph{DOSCO-2k}
For brevity, we call the 7 datasets P-Air, P-Cars, P-Ctech, P-Ins, P-Mam, P-Pets, and P-UCF, respectively. ``P'' means these datasets are created using a Places neural network. The huge ID-OOD performance gaps shown in Figure~\ref{fig:dosco_id_ood_gap} highlight the challenges of this benchmark. Following~\cite{zhai2019large}, we focus on transfer learning and create a 2k version for each dataset where the training and validation data consists of 2,000 images in total (1,600 for training and 400 for validation). Only the validation data should be used for parameter tuning. We name this benchmark \textit{DOSCO-2k}.

\section{Experiments on Image Recognition}
\label{sec:exp_im}

We evaluate our approach on the DOSCO-2k benchmark, which consists of 7 diverse datasets discussed in Section~\ref{sec:dosco}. We also conduct experiments on two commonly-used DG datasets, i.e., PACS~\cite{li2017deeper} and OfficeHome~\cite{venkateswara2017deep}. \textit{Note that this work focuses on DG without domain labels}.

\paragraph{Architecture}
We use MobileNetV3-Small~\cite{howard2019searching} as the tiny model and ResNet50~\cite{he2016deep} as the large (teacher) model for distillation-based methods. Two other tiny models specifically designed for MCUs are also evaluated: MobileNetV2-Tiny~\cite{lin2020mcunet} and MCUNet~\cite{lin2020mcunet}. The model specifications are provided in Table~\ref{tab:model_specs}.

\paragraph{Training}
The batch size is set to 32. SGD with momentum is used as the optimizer. The learning rate starts from 0.01 and decays following the cosine annealing rule. The maximum epoch is set to 100. As mentioned before, the balancing weight $\lambda$ in Eq.~\ref{eq:okd} is fixed to 0.1, which is the default setting in the KD literature~\cite{wang2021knowledge}.

\paragraph{Baselines}
We choose top-performing DG methods that do not need domain labels to compare: ERM, RSC~\cite{huang2020self}, MixStyle~\cite{zhou2021domain}, and EFDMix~\cite{zhang2022exact}. We also compare with the classic logit-based KD method~\cite{hinton2015distilling}.\footnote{Most other KD-based methods achieve similar performance with the classic KD on DOSCO-2k (see Figure~\ref{fig:overview}), so they are excluded in the experiment section for comparison.}

\paragraph{Model selection}
Model selection is a critical step when evaluating DG algorithms~\cite{gulrajani2020search}. We assume that the model can only access source data during training. Specifically, the model with the best validation performance achieved within the 100 training epochs is deployed at test data.\footnote{To better track the progress on our benchmarks, we suggest future work should conduct hyper-parameter tuning using the same model selection method, i.e., choosing parameters that give the best in-distribution validation performance.}

\begin{figure}[t]
	\centering
	\subfloat[Candidates.]{
		\includegraphics[width=.48\columnwidth]{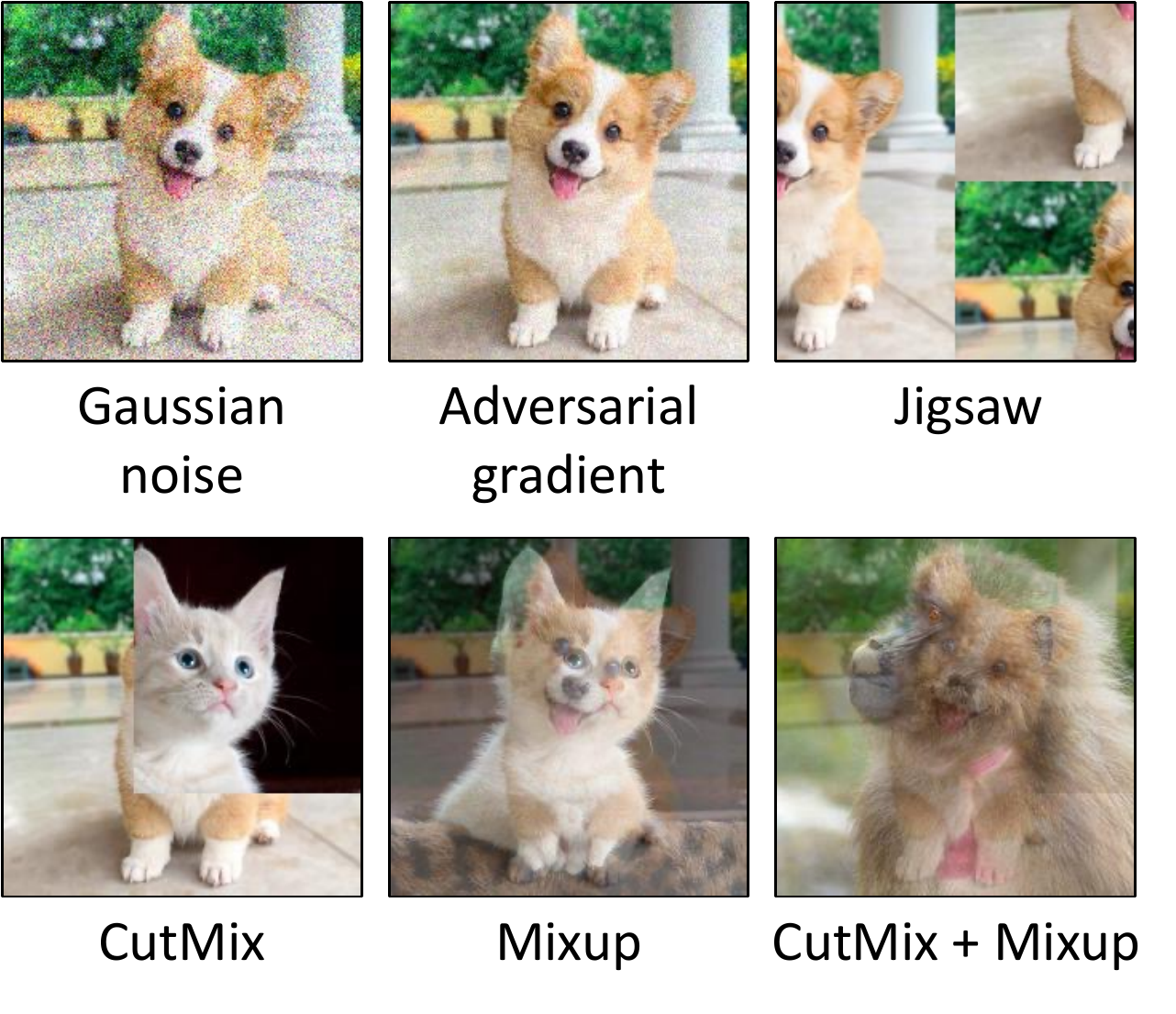}
		\label{fig:imaug}
	}
	~
	\subfloat[Performance on DOSCO-2k.]{
		\includegraphics[width=.5\columnwidth]{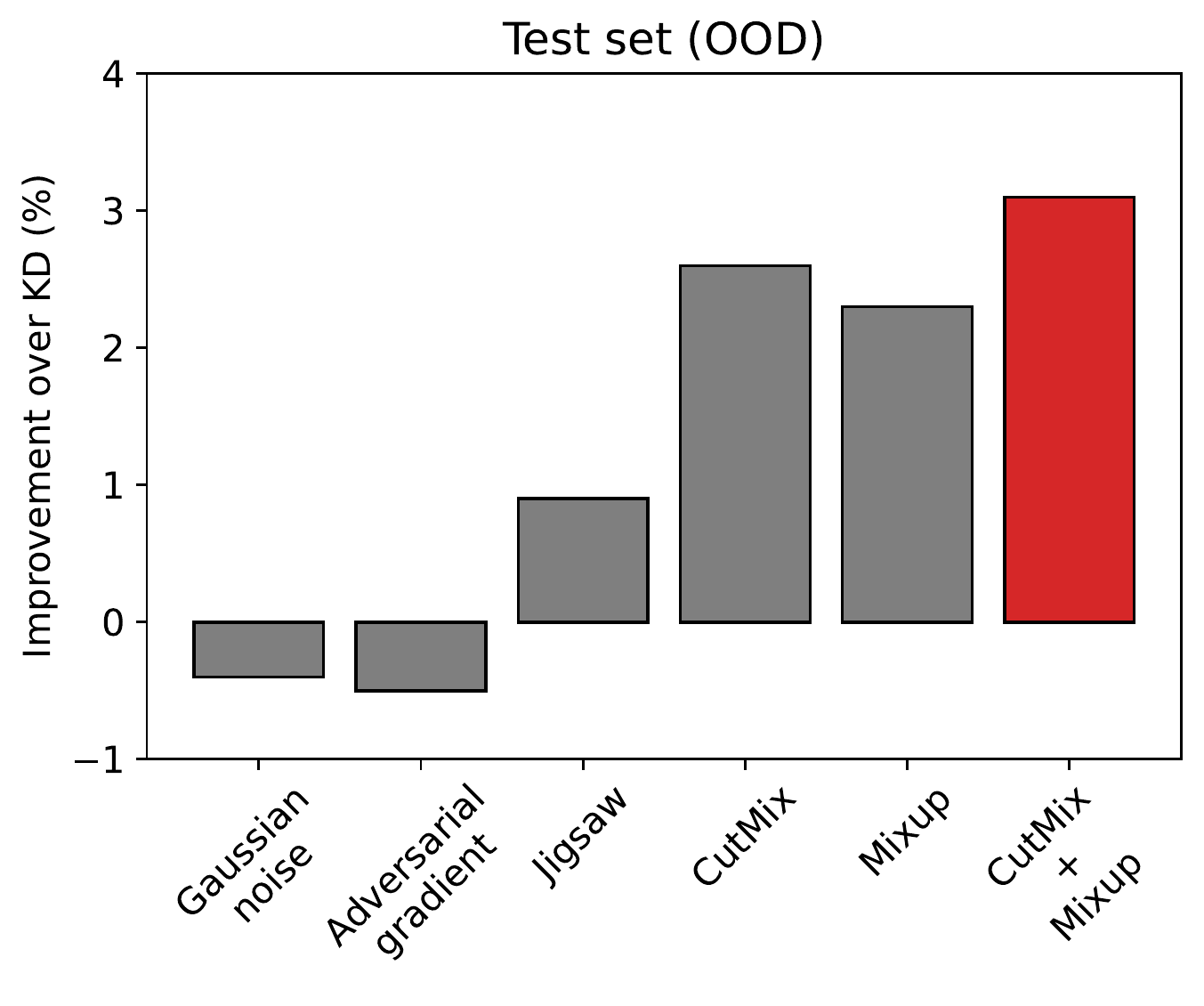}
		\label{fig:compare_imaug}
	}
	\caption{Investigation on the optimal disruptive image augmentation method for synthesizing OOD data. The combination of CutMix and Mixup performs the best for image recognition.}
\end{figure}

\begin{table*}[t]
    \tabstyle{14pt}
    \caption{
    Results on DOSCO-2k obtained with MobileNetV3-Small. The distillation-based methods substantially outperform the DG methods including the strong ERM model. OKD significantly improves upon KD with an average of 3\% increase in accuracy.
    }
    \label{tab:dosco_2k_mbv3small}
    \begin{tabu} to \textwidth {l ccccccc c}
        \toprule
        & P-Air & P-Cars & P-Ctech & P-Ins & P-Mam & P-Pets & P-UCF & \textit{Average} \\
        \midrule
        ERM & 21.3 & 18.2 & 78.6 & 38.5 & 34.4 & 67.0 & 42.2 & 42.9 \\
        RSC & 23.0 & 20.8 & 79.0 & 38.0 & 34.9 & 68.4 & 42.9 & 43.9 \\
        MixStyle & 24.6 & 22.2 & 80.9 & 37.5 & 32.3 & 67.3 & 43.7 & 44.1 \\
        EFDMix & 27.3 & 23.4 & 80.4 & 37.0 & 32.3 & 67.3 & 42.8 & 44.4 \\
        \midrule
        KD & 29.7 & 26.2 & 82.4 & 39.8 & 37.5 & 69.3 & 46.6 & 47.4 \\
        OKD & \textbf{32.1} & \textbf{30.4} & \textbf{84.4} & \textbf{42.0} & \textbf{40.8} & \textbf{73.1} & \textbf{50.4} & \textbf{50.5} \\
        \midrule
        \rowfont{\color{lightgray}}
        KD's teacher & 39.8 & 43.6 & 90.5 & 51.3 & 53.1 & 83.0 & 59.8 & 60.2 \\
        \bottomrule
    \end{tabu}
    \begin{flushleft}
        \footnotesize 
        \itshape
        Bold denotes the best result in each column (excluding the large teacher model).
    \end{flushleft}
\end{table*}

\subsection{Choosing OOD Data Generator}
The candidate methods are shown in Figure~\ref{fig:imaug}, which meet our requirement that they should push images away from the support of the data distribution but not completely disjoint---we later provide evidence that using completely disjoint data is harmful. The results of OKD using different augmentation methods are shown in Figure~\ref{fig:compare_imaug}. Most methods can bring some improvement over KD except the two noise-based methods, i.e., Gaussian noise and adversarial gradient. Notably, adversarial gradient performs the worst, suggesting that the teacher's knowledge about samples close to the decision boundary is useless---this makes sense as the teacher itself would be confused by these samples too.

CutMix and Mixup clearly stand out and the combination of them gives the best performance.\footnote{CutMix+Mixup is implemented as $\alpha \operatorname{Mixup}(\bm{x}) + (1-\alpha) \operatorname{CutMix}(\bm{x})$ where $\alpha$ is sampled from a Beta distribution. Unlike CutMix or Mixup, CutMix+Mixup may produce a mixture of three examples (see the CutMix+Mixup example in Figure~\ref{fig:imaug}).} The results suggest that the key is to keep global structure while twist local statistics. This observation would be useful for practitioners to design a better method. \textit{CutMix+Mixup is used as the default OOD data generator for image recognition tasks}

\subsection{Results on DOSCO-2k}
The full results on DOSCO-2k are reported in Table~\ref{tab:dosco_2k_mbv3small}. The first block contains DG methods, which were previously developed using large models only. In most cases, the two feature-based data augmentation methods, i.e., MixStyle and EFDMix, achieve better performance than RSC, a regularization method that mutes the most predictive subsets of neurons during training. Overall, the observations echo those reported in the literature~\cite{wiles2022fine}: (i) Not a single DG method can consistently beat ERM, e.g., none of them outperforms ERM on P-Ins; (ii) Augmentation-based methods are generally better. However, when using smaller models, the conclusions should be adjusted. When it comes to the distillation-based methods like the KD model, the margins over ERM are significant. Despite having a simple design, OKD outperforms KD on all datasets with an average improvement of 3\%, which strongly demonstrates its potential for solving the on-device DG problem. The results also serve as an ablation study for justifying the design of OKD: the only difference between OKD and KD is the proposed distillation loss computed on synthetic OOD data (see Eq.~\ref{eq:okd}).

\begin{figure}[t]
    \centering
    \subfloat[MobileNetV2-Tiny.]{
        \includegraphics[width=.48\columnwidth]{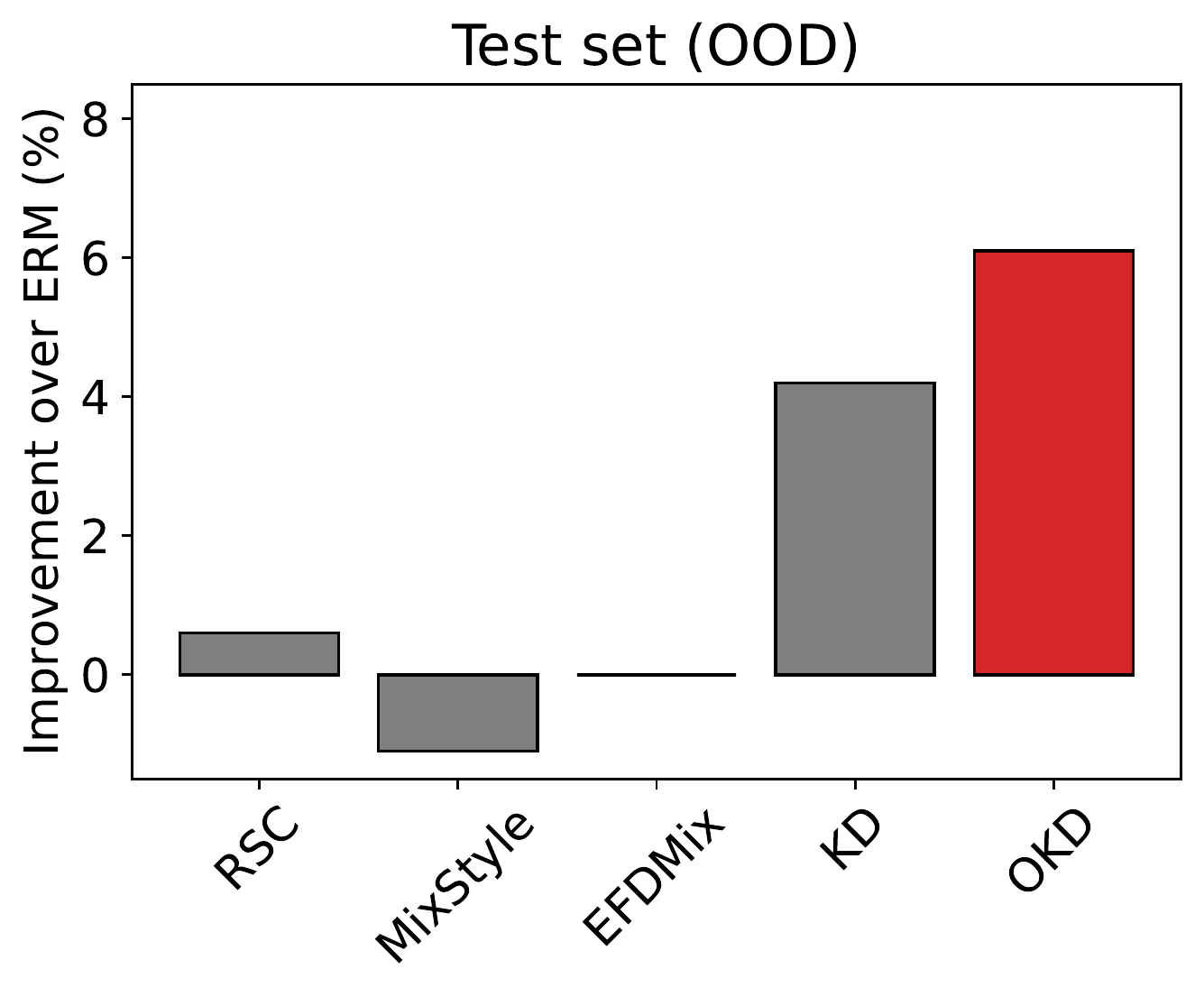}
    }
    ~
    \subfloat[MCUNet.]{
        \includegraphics[width=.48\columnwidth]{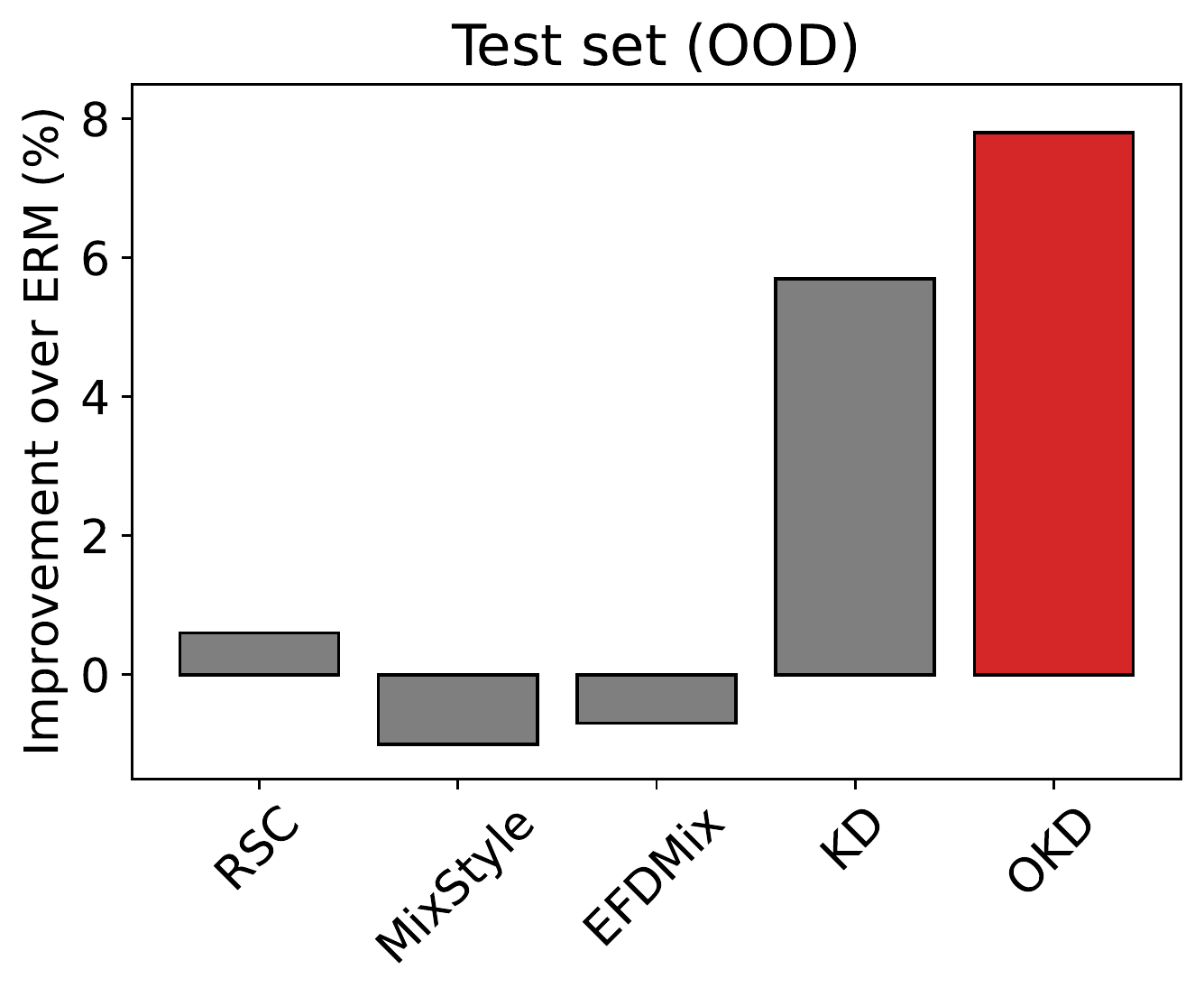}
    }
    \caption{Results of other tiny architectures. OKD still performs the best and the margins are significant.}
    \label{fig:mbv2tiny_mcunet_vs_erm}
\end{figure}

\paragraph{Results of other tiny architectures}
We further evaluate our approach using two other tiny neural networks, i.e., MobileNetV2-Tiny and MCUNet, which are specifically designed for MCUs~\cite{lin2020mcunet}. As shown in Table~\ref{tab:model_specs}, these two architectures are half the size of MobileNetV3-Small, meaning that their capacity is further shrunk down, and as a consequence, improving their DG performance would be much more challenging. We repeat the same experiments using these two architectures on DOSCO-2k. The results are summarized in Figure~\ref{fig:mbv2tiny_mcunet_vs_erm}. Interestingly, the two feature-based data augmentation methods, which could beat ERM and RSC when MobileNetV3-Small is used, are no longer competitive in this setting. RSC, on the other hand, is able to gain some improvement over ERM, but the gains are only marginal (less than 1\%) and much smaller than those obtained with MobileNetV3-Small (cf.~Figure~\ref{fig:teaser_mbv3small}). Our approach OKD still maintains its dominance and achieves nontrivial improvements over the strong KD model for both architectures.

\begin{SCfigure*}
    \centering
    \includegraphics[width=.8\textwidth]{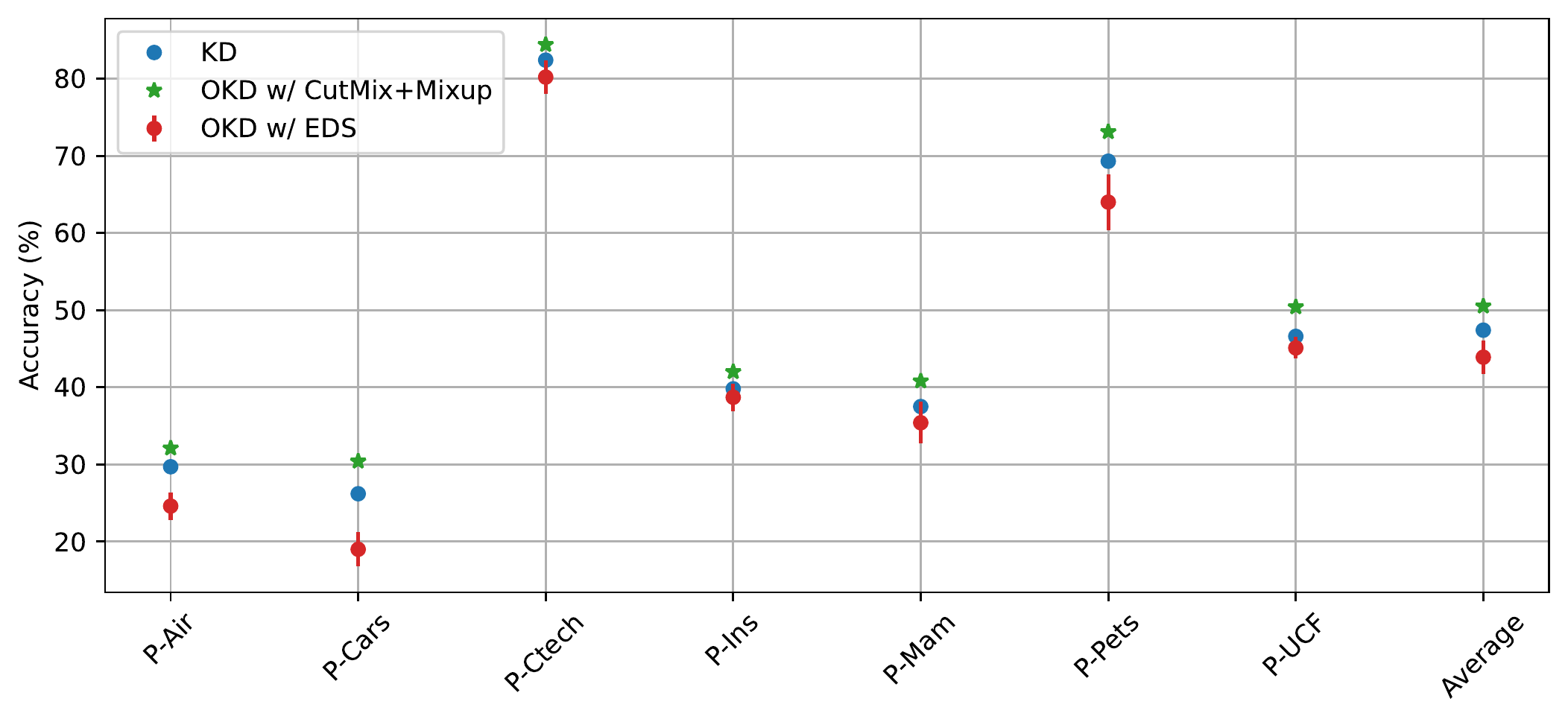}
    \caption{Image transformation vs.~external data source (EDS). The vertical bars of OKD w/ EDS correspond to standard deviations. Clearly, EDS does not work and the performance is even worse than KD, which confirms our assumption that the OOD data should not be completely disjoint from the source data.}
    \label{fig:aug_vs_eds}
\end{SCfigure*}

\begin{table}[t]
    \tabstyle{16pt}
    \caption{Ablation study of OKD w/ Jigsaw. Shuffling more patches leads to worse results, which emphasizes the importance of maintaining global structure in image statistics.}
    \label{tab:ablation_jigsaw}
    \begin{tabular}{l ccc}
        \toprule
        & $k\!=\!4$ & $k\!=\!16$ & $k\!=\!64$ \\
        \midrule
        Accuracy & 48.3 & 46.9 & 40.5 \\
        \bottomrule
    \end{tabular}
    \begin{flushleft}
        \footnotesize
        \itshape
        $k$ denotes the total number of patches to shuffle.
    \end{flushleft}
\end{table}

\paragraph{Image transformation vs.~external data source}
We have mentioned earlier that the key to making OKD work is to not produce OOD samples that are completely disjoint from the data distribution. To validate this design choice, we conduct large-scale experiments by training OKD in the following way: For each dataset on DOSCO-2k, we use a different dataset as the OOD data source, repeat such an experiment for all available OOD data sources, and average the results. Figure~\ref{fig:aug_vs_eds} confirms our assumption as the results of using external data source are much worse than those using the image transformation. We further verify this assumption using the Jigsaw version of OKD where the number of patches to shuffle is increased from 4 to 16 to 64. Table~\ref{tab:ablation_jigsaw} shows that the more patches we shuffle (the farther away the synthetic data is from the training data distribution), the worse the performance.

\begin{table}[t]
	\tabstyle{15pt}
	\caption{Results on PACS and OfficeHome.}
	\label{tab:pacs_oh_mbv3small}
	\begin{tabu} to \textwidth {l cc}
		\toprule
		& PACS & OfficeHome \\
		\midrule
		ERM & 72.3 & 53.9 \\
		RSC & 71.5 & 53.2 \\
		MixStyle & 73.0 & 53.9 \\
		EFDMix & 74.7 & 53.8 \\
		\midrule
		KD & 73.6 & 56.2 \\
		OKD & \textbf{76.8} & \textbf{59.8} \\
		\midrule
		\rowfont{\color{lightgray}}
		KD's teacher & 82.1 & 64.9 \\
		\bottomrule
	\end{tabu}
\end{table}

\subsection{Results on PACS and OfficeHome}
Table~\ref{tab:pacs_oh_mbv3small} shows the results on PACS and OfficeHome, each averaged over all test domains. Overall, the observations are similar to those on DOSCO-2k: distillation works better; OKD significantly reduces the gaps with the teacher.

\section{Experiments on Speech Recognition}
\label{sec:exp_speech}

\begin{table}[t]
	\centering
	\caption{Results on Google Speech Commands where the training and test speakers and different. OKD clearly outperforms KD and ERM.}
	\subfloat[Results of OKD with different augmentation methods.]{
		\tabstyle{13pt}
		\begin{tabular}{l c}
			\toprule
			& Acc \\
			\midrule
			Mixup & 65.9 \\
			Noise & 66.7 \\
			Mask & \textbf{67.1} \\
			\bottomrule	
		\end{tabular}
		\label{tab:speech_aug}
	}
	\hspace{2em}
	\subfloat[Comparison between ERM, KD and OKD (using Mask).]{
		\tabstyle{13pt}
		\begin{tabular}{l c}
			\toprule
			& Acc \\
			\midrule
			ERM & 56.3 \\
			KD & 64.8 \\
			OKD & \textbf{67.1} \\
			\bottomrule	
		\end{tabular}
		\label{tab:speech_compare}
	}
\end{table}

For speech recognition, we choose Google Speech Commands, which is a dataset of 35 commands spoken by different people. Each sample is a 1-second long audio track. We build a DG version of this dataset by splitting the data into training, validation and test sets with non-overlapping speakers, following a $7\!:\!1\!:\!2$ ratio.

\paragraph{Architecture}
We use CNNs to directly process raw waveforms. Specifically, M3~\cite{dai2017very}, a 3-layer CNN with 0.2M parameters, is used as the tiny model while M11~\cite{dai2017very}, a 11-layer CNN with 1.8M parameters, is used as the teacher model, particularly for KD and OKD.

\paragraph{Training}
Adam~\cite{kingma2014adam} is used as the optimizer. The batch size is 256. The learning rate is 0.01, which is decayed using the cosine annealing rule. The maximum epoch is 60. The model selection strategy follows that in the previous experiments.

\subsection{Results on Google Speech Commands}
We first study which augmentation method best suits OKD in speech recognition. Three candidates are evaluated: (1) Mixup, which randomly mixes two waveforms; (2) Noise, which adds random Gaussian noise to waveforms; (3) Mask, which randomly zeros out a small portion of waveforms. As shown in Table~\ref{tab:speech_aug}, Mask gives the best performance. Next, we compare OKD with ERM and KD. The results are shown in Table~\ref{tab:speech_compare}. KD substantially improves upon ERM by about 9\%. Again, OKD further boosts the performance by about 3\%, which is significant. It is worth mentioning that the teacher model achieves 91.0\% accuracy, which means the gap is yet closed.

\begin{table}[t]
	\tabstyle{10pt}
	\caption{Ablation study on data augmentation. KD+Aug uses the same augmentation method as OKD. Clearly, the improvement of OKD over KD comes from the proposed distillation loss instead of the effect of data augmentation.}
	\label{tab:ablation_aug}
	\begin{tabular}{lcc}
		\toprule
		& DOSCO-2k & Speech \\
		\midrule
		KD & 47.4 & 64.8 \\
		KD+Aug & 48.7 & 65.6 \\
		OKD & \textbf{50.5} & \textbf{67.1} \\
		\bottomrule
	\end{tabular}
\end{table}

\paragraph{Effect of data augmentation}
To understand whether OKD's improvement over KD comes from the proposed distillation loss or just the effect of data augmentation, we compare with a variant of KD (called KD+Aug) where the distillation term is kept unchanged but the cross-entropy term is computed on augmented data (the same augmentation method used by OKD is applied to KD+Aug). Table~\ref{tab:ablation_aug} shows the results on both image and speech recognition, which effectively clear the doubt that the improvement might come from data augmentation.

\section{Related Work}
We briefly review two areas closely related to our research, namely domain generalization (DG) and knowledge distillation (KD). See~\cite{zhou2022domain,wang2021knowledge} for more comprehensive surveys in these two areas.

\paragraph{Domain generalization}
The majority of DG methods can be grouped into three categories: domain alignment, meta-learning, and data augmentation. Domain alignment methods often employ a distance measure like Maximum Mean Discrepancy~\cite{li2018domain} or adversarial learning~\cite{zhang2022delving} to reduce the feature distributions between two or multiple source domains. Meta-learning methods, on the other hand, adopt the notion of learning-to-learn and typically perform model learning using pseudo-source and pseudo-target data~\cite{li2018learning,balaji2018metareg,dou2019domain,shi2022gradient}. Data augmentation methods aim to diversify the training data, which is often achieved by learning a generative model~\cite{zhou2020learning,zhou2020deep} or mixing data at the input~\cite{xu2020adversarial,yao2022improving} or feature-level~\cite{zhou2021domain,zhang2022exact}. More recent research has explored test-time adaptation~\cite{wang2020tent,zhang2021memo,iwasawa2021test}, which updates the model on-the-fly using a test datapoint or minibatch, and multimodal learning, such as learning a joint embedding space for image and language~\cite{min2022grounding}.

Our research significantly differs from existing ones in that we study DG in the context of tinyML and for the first time present a systematic study on methods that can improve DG for tiny neural networks. Existing DG methods are mainly developed using large models so it was unclear whether they can be applied to tiny models. Our research provides a timely answer to this question.

\paragraph{Knowledge distillation}
KD is a popular technique used in model compression~\cite{bucilua2006model} as well as other areas like semi-supervised learning~\cite{chen2020big}. The most basic form of KD is to minimize the KL divergence between the student and teacher's outputs~\cite{hinton2015distilling} (reviewed in Section~\ref{sec:approach;subsec:kd}). Several follow-ups extended KD by re-designing representations (to be transferred) based on, e.g., hints~\cite{romero2014fitnets}, probabilistic distributions~\cite{passalis2018learning}, attention maps~\cite{zagoruyko2017paying}, and mutual information~\cite{ahn2019variational}. Some variants sought other useful knowledge sources like inter-instance correlations~\cite{tian2020contrastive,park2019relational,tung2019similarity} or self-supervision~\cite{xu2020knowledge}.

More related to our work are data-free KD methods~\cite{fang2021mosaicking,binici2022preventing}, which typically use generative modeling to synthesize training data. Differently, our work addresses on-device DG, a new problem that can be viewed as a unification of the areas of DG and KD. Our results unveil an interesting finding that has not been brought up in the KD literature: the teacher-student gap on OOD data is larger than that on ID data, which leads to concerns over mobile DG applications.

\section{Conclusion, Limitation and Future Work}

This paper is the first to investigate how to improve DG for tiny neural networks. The results reveal that current state-of-the-art DG methods, which were developed using large models, do not work well in our on-device DG setting and perform inconsistently for different tiny network architectures. For example, MixStyle and EFDMix can improve upon ERM when using MobileNetV3-Small but their performance plunges below ERM's when using two much smaller architectures specifically designed for MCUs.

Overall, the results strongly suggest that tiny neural networks with small capacity and low complexity should be trained differently than their large counterparts for DG applications. The proposed OKD approach, which simply adds a new distillation loss term to KD and does not introduce any extra parameter to the model, demonstrates great potential for solving the problem. OKD can serve as a strong baseline to build for future work.

Although OKD's improvements are significant, the performance gap with large models is still huge: about 10\% difference on DOSCO-2k and 24\% on speech recognition. Therefore, on-device DG is yet to be solved. An obvious limitation of OKD is that training needs to be performed on devices with sufficient compute and memory---since distillation requires a large model---rather than on low-power devices directly. It would be interesting to explore ways that enable full on-device learning. Many other interesting questions are also worth exploring: Can we design more advanced OOD data generators or make the function fully-learnable? Is it possible to combine KD with DG methods while pursuing efficacy? How about designing parameter-efficient modules or learning strategies that can offer a good trade-off between performance and efficiency? To name a few.

\paragraph{Broader impact}
Scaling up neural networks has often been used to boost performance but the extra cost---both economically and environmentally---it brings is also nontrivial and should not be ignored. Our research aims to bring down the deployment cost of AI applications by making tiny models more generalizable, hence more comparable to their large counterparts. In doing so, we promote the use of smaller, more economic models that consume less energy and require less compute. Our study unifies the areas of domain generalization and model compression, and provides timely insights to both communities.

{\small
\bibliographystyle{ieee_fullname}
\bibliography{egbib}
}

\end{document}